\documentclass{article}

\usepackage{microtype}
\usepackage{graphicx}
\usepackage{booktabs} 
\usepackage{comment}
\usepackage{adjustbox}
\usepackage{caption}
\usepackage{subcaption}

\usepackage{hyperref}

\usepackage[accepted]{icml2024}

\usepackage{amsmath}
\usepackage{amssymb}
\usepackage{mathtools}
\usepackage{amsthm}

\usepackage[capitalize,noabbrev]{cleveref}

\theoremstyle{plain}
\newtheorem{theorem}{Theorem}[section]
\newtheorem{proposition}[theorem]{Proposition}
\newtheorem{lemma}{Lemma}
\newtheorem{corollary}[theorem]{Corollary}
\theoremstyle{definition}
\newtheorem{definition}[theorem]{Definition}

\theoremstyle{remark}
\newtheorem{remark}[theorem]{Remark}
\newcommand{\R}{\mathbb{R}}
\newcommand{\F}{\mathcal{F}}
\newcommand{\veF}{\mathcal{F}_{v \unlhd e}}
\newcommand{\ueF}{\mathcal{F}_{u \unlhd e}}

\usepackage[textsize=tiny]{todonotes}

\newcommand{\framedtext}[1]{%
\par%
\noindent\fbox{%
    \parbox{\dimexpr\linewidth-2\fboxsep-2\fboxrule}{#1}%
}%
}

\begin{document}

\twocolumn[
\icmltitle{Joint Diffusion Processes as an Inductive Bias in Sheaf Neural Networks}

\begin{icmlauthorlist}
\icmlauthor{Ferran Hernandez Caralt}{UPC-Ferran}
\icmlauthor{Guillermo Bernárdez Gil}{UCSB}
\icmlauthor{Iulia Duta}{UniCamb}
\icmlauthor{Pietro Liò}{UniCamb}
\icmlauthor{Eduard Alarcón Cot}{UPC-Eduard}
\end{icmlauthorlist}

\icmlaffiliation{UPC-Ferran}{FME, FIB and ETSETB, Universitat Politècnica de Catalunya-BarcelonaTech, Barcelona, Spain}
\icmlaffiliation{UPC-Eduard}{Department of Electronic Engineering, Universitat Politècnica de Catalunya-BarcelonaTech, Spain, Barcelona}
\icmlaffiliation{UniCamb}{Department of Computer Science and Technology, University of Cambridge, Cambridge, United Kingdom}
\icmlaffiliation{UCSB}{Department of Electrical and Computer Engineering, UC Santa Barbara, Santa Barbara, United States of America}


\icmlsetsymbol{equal}{*}

\icmlcorrespondingauthor{Ferran Hernandez Caralt}{ferran.hernandez.caralt@estudiantat.upc.edu, ferranhernandezc@gmail.com}
\icmlkeywords{Machine Learning, ICML}

\vskip 0.3in
]



\printAffiliationsAndNotice{} 

\begin{abstract}
    Sheaf Neural Networks (SNNs) naturally extend Graph Neural Networks (GNNs) by endowing a cellular sheaf over the graph, equipping nodes and edges with vector spaces and defining linear mappings between them. While the attached geometric structure has proven to be useful in analyzing heterophily and oversmoothing, so far the methods by which the sheaf is computed do not always guarantee a good performance in such settings. 
    In this work, drawing inspiration from opinion dynamics concepts, 
    we propose two novel sheaf learning approaches that \textit{(i)} provide a more intuitive understanding of the involved structure maps, \textit{(ii)} introduce a useful inductive bias for heterophily and oversmoothing, and \textit{(iii)} infer the sheaf in a way that does not scale with the number of features, thus using fewer learnable parameters than existing methods. 
    In our evaluation, we show the limitations of the real-world benchmarks used so far on SNNs, and design a new synthetic task --leveraging the symmetries of $n$-dimensional ellipsoids-- that enables us to better assess the strengths and weaknesses of sheaf-based models. 
    Our extensive experimentation on these novel datasets reveals valuable insights into the scenarios and contexts where SNNs in general --and our proposed approaches in particular-- can be beneficial.
\end{abstract}

\section{Introduction}
Graph Neural Networks (GNNs)~\cite{scarselli2008graph,kipf2016semi,velivckovic2017graph} have become very popular as a way to model and process relational data, demonstrating remarkable performance in a wide range of applications and tasks~\cite{zhou2020graph,qiu2018deepinf,rusek2019unveiling}. Nonetheless, two main problems frequently appear when dealing with GNNs: they perform poorly on heterophilic settings~\cite{zhu2020beyond,luan2022revisiting}, and exhibit over-smoothing behaviour~\cite{nt2019revisiting,oono2019graph}. The first problem emerges because most models implicitly assume graph homophily (i.e. edges are expected to connect similar nodes), whereas the second one has to do with the tendency of deep GNNs to produce features too uniform to be useful.

Sheaf Neural Networks (SNNs)~\cite{hansen2020sheaf}, originally designed as a natural generalization of GNNs with a \textit{possibly} non-trivial underlying graph "geometry", have been proven a powerful tool to analyze the aforementioned issues \cite{bodnar2022neural}. In particular, SNNs endow a (cellular) sheaf~\cite{curry2014sheaves} 
over the graph, equipping each node and edge with a vector space and defining a linear application between these spaces for each incident edge-node. The particular sheaf choice is reflected in the graph Laplacian \cite{hansen2019toward} operator, the properties of the corresponding diffusion equation, and the convolutional models that discretise the equation \cite{bodnar2022neural}. At first, the sheaf was defined through domain knowledge \cite{hansen2020sheaf}, but recent papers have proposed new ways to work with this structure, such as learning the sheaf using a learnable parametric function \cite{bodnar2022neural}, inferring the graph connection Laplacian directly from data at preprocessing time \cite{barbero2022sheaf}, using the wave equation on sheaves to design the model \cite{suk2022surf}, or even introducing non-linearities in the process \cite{zaghen2024nonlinear}. However, the sheaf structure is typically inferred through universal approximators (e.g. via a Multi-Layer Perceptron (MLP)) that do not intrinsically incorporate an inductive bias for heterophilic data.

\paragraph{Contributions:} 
Motivated by this fact, in this work we propose new SNN variants specifically \textbf{tailored for learning on heterophilic data}, and which learn the sheaf \textbf{explicitly} to prevent oversmoothing. Drawing inspiration from \cite{hansen2021opinion} and some of its opinion dynamics-based interpretations of sheaves, we introduce a SNN formulation whose theoretical guarantees of class separation \textbf{no longer depend on learnable parameters}. 
As a practical result of this, our proposed models are able to compute the sheaf with a significantly lower number of parameters compared to existing SNN models. 
Lastly, and in order to overcome the current limitations of SNN benchmarks, we also propose a \textbf{novel synthetic framework} to evaluate the performance of sheaf-based approaches on node classification tasks. In this new framework, we leverage symmetries of the surface of n-dimensional ellipsoids to create non-noisy distinguishable classes \textbf{where, even under high homophily, GCNs suffer from oversmoothing}, thus making the separation of classes a hard problem for regular GNNs. In this new setting, we show how the different approaches to learn the sheaf may be advantageous or disadvantageous depending on the characteristics of the data.

\section{Background}
Let us begin contextualizing SNNs by briefly revisiting one of the most popular GNNs: the Graph Convolutional Network (GCN) \cite{kipf2016semi}. The $t$ layer of a GCN is typically formulated as
\begin{equation}
    X(t) = X(t-1)-\sigma(\hat{D}^{-\frac{1}{2}}L\hat{D}^{-\frac{1}{2}}X(t-1)W(t)),
\end{equation}
where $W(t)$ and $X(t)$ represent the weights and node features at step $t$, $\sigma$ a non-linear activation function, $L = I - A$ the graph Laplacian, and $\hat{D}$ the node degree matrix of $A+I$. We note that this can be seen as using an MLP on top of the following Ordinary Differential Equation (ODE) over the original node signal:
\begin{equation}
\label{eq:RegGraphDiff}
    \frac{d}{dt}X = -LX,
\end{equation}
which in turn satisfies the following property:
\begin{proposition}
\label{prop:GrafSmooth}
    Let $G=(V,E)$ be a graph with an associated graph Laplacian $L$ and $X(t)$ a node signal satisfying \eqref{eq:RegGraphDiff}, then $lim_{t \rightarrow \infty} X(t)$ is constant on all connected components.
    
\textit{Proof in Appendix \ref{appendix:proof_oversmoothing}.}
\end{proposition}

This behaviour, root of the so-called \textbf{oversmoothing} phenomena, is a common issue when handling graph classification tasks with multiple-layered GNNs~\cite{chen2020measuring}, preventing them from building discriminative representations for different classes. 
In particular, since the ODE-based diffusion process is locally defined through graph connectivity, oversmoothing problems are even more compelling when dealing with heterophilic data~\cite{yan2022two} (i.e. graphs where nodes from different classes tend to be connected).
To address this particular context, the authors of \cite{bodnar2022neural} proposed to learn a sheaf-structure on the graph. Before reviewing the results of this work, though, let us first introduce the concept of a cellular sheaf and some useful notation:
\begin{definition}
    Given an undirected graph $G = (V,E)$, we will call $(G,\mathcal{F})$ a cellular sheaf when $\mathcal{F}$ 
    \begin{itemize}
        \item assigns, for each $v \in V$ and $e \in E$, the corresponding vector spaces $\mathcal{F}(v)$ and $\mathcal{F}(e)$, which are called \textit{stalks};
        \item defines a linear map  $\mathcal{F}_{v \unlhd e}: \mathcal{F}(v) \rightarrow \mathcal{F}(e)$ for each incident $v \unlhd e$ node-edge pair, known as \textit{restriction maps}.
    \end{itemize}
\end{definition} 

\begin{definition}
    We will call the direct sum of all the node vector spaces the space of \textit{0-cochains}, $C^0(G;\F) = \bigoplus_{v \in V} \F(v)$. Analogously we can define the space of \textit{1-cochains}, $C^1(G;\F) = \bigoplus_{e \in E} \F(e)$. $x_u$ and $x_e$ will be the projections unto the spaces $\F(u),\F(e)$.
\end{definition} 

\begin{definition}
    We will call the space of global sections the subspace $H^0(G;\F)$$ \subseteq C^0(G;\F)$ that satisfies $$H^0(G;\F) = \{x \in C^0(G;\F) | \ueF x_u = \veF x_v, \text{ } \forall e \in E\}.$$
\end{definition} 

From what we've seen so far we can define a new linear map in the following way:

\begin{definition}
    We define the coboundary map $\delta : C^0(G;\F) \rightarrow C^1(G;\F)$ where\footnote{Note that an arbitrary orientation of the edges 
    must be defined. However, the particular choice is irrelevant for our purposes.}  $$(\delta x)_e = \ueF x_u - \veF x_v.$$
\end{definition} 

Now, with this extra structure, a new Laplacian may be defined in the following way:
\begin{definition}
    We define the \textit{sheaf laplacian} $\Delta_\F : C^0(G;\F) \rightarrow C^0(G;\F)$ as $\Delta_\F = \delta^T\delta$. Given a node signal $X(t)$ on a graph $G = (V,E)$ with an associated sheaf $\F$, then its sheaf Laplacian applied to node $u$ would be:
    \begin{equation}
        (\Delta_\F x)_u = \sum_{u,v \unlhd e} \ueF^T(\ueF x_u - \veF x_v).
    \end{equation}
\end{definition} 

This Laplacian can be used like the graph Laplacian in \eqref{eq:RegGraphDiff} to define an ODE over $X \in C^0(G;\F)$:
\begin{equation}
\label{eq:SheafDiffusion}
    \frac{d}{dt}X = -\Delta_{\mathcal{F}}X,
\end{equation}
resulting in a generalization of GCNs when discretised and used as a diffusion operator to define SNNs:

\begin{figure*}[]
\begin{subfigure}[c]{0.33\textwidth}
\centering
    \includegraphics[scale=0.15]{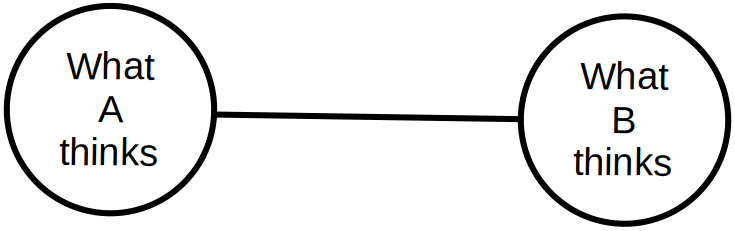}
\caption{\label{subfig:SheafMotivA}}
\end{subfigure}
\begin{subfigure}[c]{0.33\textwidth}
    \centering
    \includegraphics[scale=0.15]{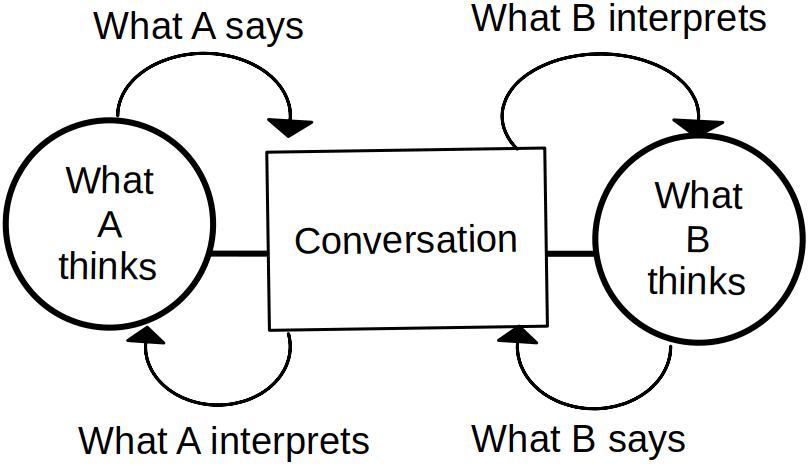}
\caption{\label{subfig:SheafMotivB}}
\end{subfigure}
\begin{subfigure}[c]{0.33\textwidth}
    \centering
    \includegraphics[scale=0.15]{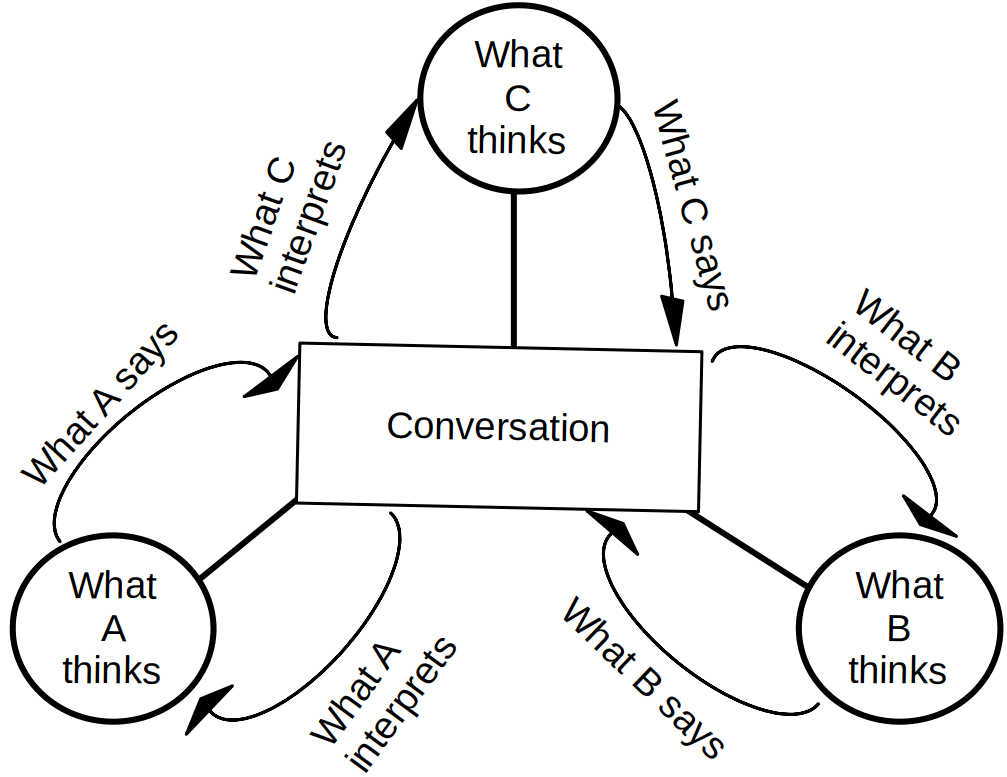}
    \caption{\label{subfig:SheafMotivC}}
\end{subfigure}
   \caption{Visual representations of (\subref{subfig:SheafMotivA}) a pair-wise interaction modelled by a graph, (\subref{subfig:SheafMotivB}) a pair-wise interaction modelled by a sheaf over a graph, and (\subref{subfig:SheafMotivC}) a higher-order interaction modelled by a sheaf over a hypergraph.}
\label{fig:SheafMotiv}
\end{figure*}
\begin{definition}
\label{def:SNN}
    Let us denote $X(t) \in \R^{nd} \times \R^f$, where $n$ is the number of nodes, $d$ is the dimension of the stalks, $f$ the number of feature channels, and $t$ the layer of the neural network. The update equation that corresponds to the $t$ layer of a SNN is:
    \begin{equation}
        X(t+1) = X(t) - \sigma(\Delta_{\F(t)}(I_n \otimes W_1(t)) X(t) W_2(t))
    \end{equation}
    where the restriction maps of $\F(t)$ are computed like \cite{bodnar2022neural} --i.e. $\ueF(t) = MLP(x_u||x_v)$ with $v \unlhd e$--, and the matrices $W_1(t), W_2(t)$ consist of learnable parameters that act like convolutions.
\end{definition}

The key point of this SNN is to learn a Laplacian with a richer set of equilibrium points that is no longer limited by a proposition like \ref{prop:GrafSmooth}, thus avoiding oversmoothing. In this regard, we have the following result:
\begin{proposition}
\label{prop:HodgeTheorem}
    \cite{bodnar2022neural} Let $G$ be a graph with an associated sheaf $\F$ and a node signal $X(t)$. If $X(t)$ satisfies (\ref{eq:SheafDiffusion}), then $lim_{t \rightarrow \infty} X(t) \in H^0(G;\F)$.
\end{proposition}

In particular, this means that instead of converging to a configuration where $x_u = x_v$ for all nodes $u,v$, we now converge to a configuration where $\ueF x_u = \veF x_v$. So, as intended, by considering the right restriction maps we should be able to have different classes be connected in our graph WITH different features. It is important to note, however, that unless the maps are orthogonal we have no guarantees that $H^0(G;\F) \neq \{0\}$ \cite{bodnar2022neural}.


\begin{remark} \label{remark:restriction_learning}
    In order to learn these restriction maps, existing SNNs typically set $\ueF = MLP(x_u||x_v)$, getting a sheaf universal approximation result \cite{bodnar2022neural}. Nevertheless, this choice does not introduce any inductive bias to ensure that learnt maps are able to deal with heterophily and oversmoothing. 
\end{remark}

\section{Opinion Dynamics Inspired Sheaf Neural Networks}

Addressing the point raised in Remark \ref{remark:restriction_learning}, this section explores new ways of predicting the sheaf with a stronger inductive bias towards heterophilic data. To do so, we turn to opinion dynamics --the field that studies how opinions in a group of people spread--~\cite{xia2011opinion} and put our focus on some ODEs proposed in \cite{hansen2021opinion} to analyze opinions' exchanges from a cellular sheaf perspective.

\subsection{Motivation}
\begin{remark}
    If we encode a social network as a simple graph, the edges can only represent if two nodes interact or if they don't interact.
\end{remark}

With this remark, it is straightforward to see that a naive graph-based modeling of opinion dynamics would prevent us from capturing the different types of interactions between different types of opinions.
In other words, using a graph to spread opinions represents a myopic way of handling the complexity of human communication. Instead of blindly exchanging their private opinions (Fig. \ref{subfig:SheafMotivA}), humans engage in conversations, where they have the freedom to express different nuances of their beliefs and draw their own conclusions based on how the conversation goes (Fig. \ref{subfig:SheafMotivB}). Precisely, cellular sheaves offer us the necessary tools to formalise this, 
providing with an intuitive interpretation of the sheaf~\cite{hansen2021opinion}: 
$x_u$ now represents the node's private opinion, while $\ueF x_u$ is the public opinion it decides to share with node $v \unlhd e$. It is worth noting that the same rationale can be applied to hypergraphs, having more than 2 people participate in a conversation (Fig. \ref{subfig:SheafMotivC}). 

Bearing this in mind, we observe that Graph Diffusion (\ref{eq:RegGraphDiff}) is a process that makes the node's true private opinions agree, while in Sheaf Diffusion 
(\ref{eq:SheafDiffusion}) the agreement is achieved at the level of node's public opinions. 
This means that current variants of SNNs assume \textit{(i)} a fixed communication space, and \textit{(ii)} a diffusion process that modifies the private opinion to reach a consensus in that space. However, a more natural approach would be to allow the communication channel to evolve simultaneously, which is also important to guarantee the convergence to non-trivial agreements. As we know from Proposition \ref{prop:HodgeTheorem}, Eq. \eqref{eq:SheafDiffusion} must converge to a point satisfying a system of linear equations, but in general, this system might not have non-zero solutions. In this regard, we recall that \cite{bodnar2022neural} only proved the existence of non-trivial solutions when considering orthogonal restriction maps.

\subsection{Learning to Lie}
In our aim to model more intricate interactions, we turn to the ODE introduced and studied in \cite{hansen2021opinion} as \textit{Learning to Lie}. With this diffusion, we may overcome the aforementioned limitations by making the restriction maps evolve to fit the opinions instead of the other way around:
\begin{equation} \label{eq:learning_to_lie}
 \frac{d}{dt} \mathcal{F}_{u \unlhd e} = -(\mathcal{F}_{u \unlhd e}x_u - \mathcal{F}_{v \unlhd e}x_v)x_u^T.
\end{equation}

\begin{figure}
    \centering
    \includegraphics[scale=0.15]{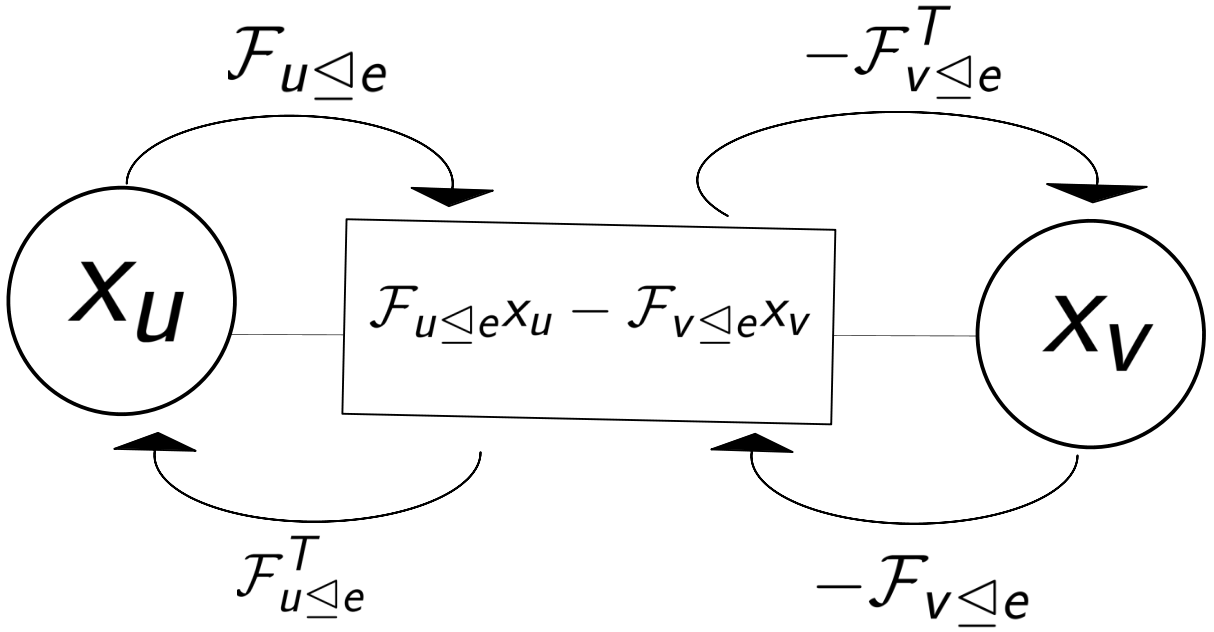}

    \includegraphics[scale=0.15]{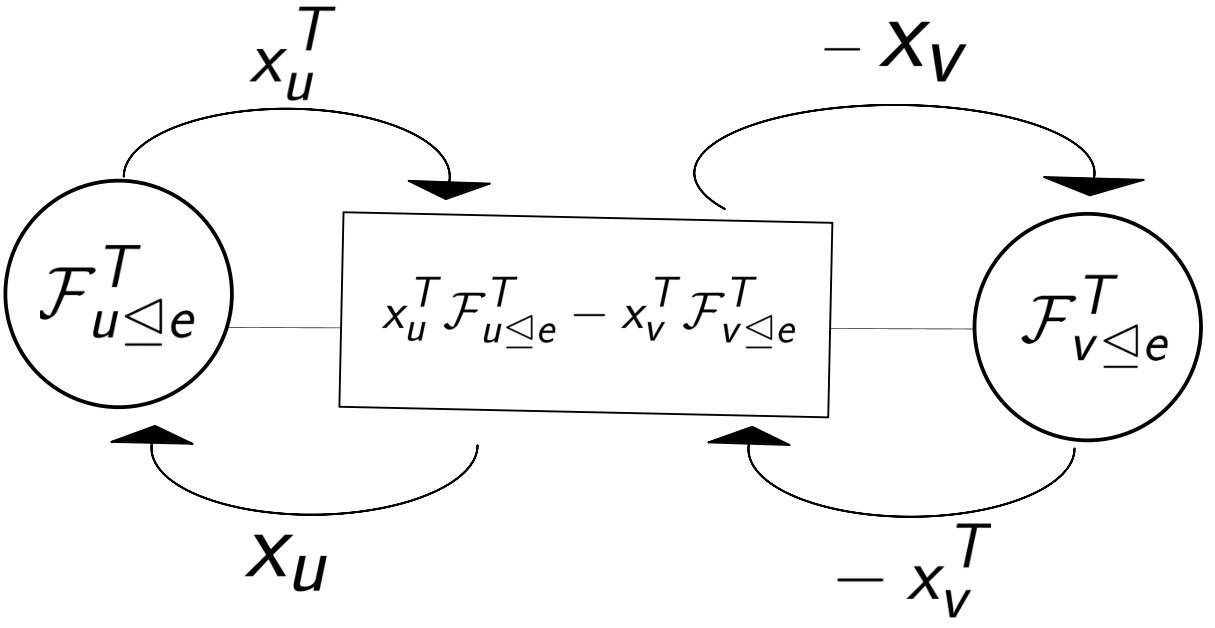}
    \caption{Top: a diagram representing Regular Sheaf Diffusion. Bottom: a representation of the \textit{Learning to Lie} ODE diffusion.}
    \label{fig:DiagramaDualSheaves}
\end{figure}

Note that this ODE can be seen as a "dual" version of sheaf diffusion where we use the features $x_u$ as maps by transposing them to diffuse the restriction maps; we can slightly rewrite Eq. \eqref{eq:learning_to_lie} as
\begin{equation}
\frac{d}{dt} \mathcal{F}_{v \unlhd e}^T = -x_u(x_u^T\mathcal{F}_{u \unlhd e}^T - x_v^T\mathcal{F}_{v \unlhd e}^T)
\end{equation}
for a more explicit analogy (please see Figure \ref{fig:DiagramaDualSheaves} for a visual representation). 
In fact, given a sheaf Laplacian defined with the $X$ signal, and denoting by $\F^*$ a matrix with all the restriction maps transposed, we can formulate this Dual Diffusion Process using the standard notation
\begin{equation}
\label{eq:DualDiff}
\frac{d}{dt}\F^* = -\Delta_{X}\F^*.
\end{equation}

As proven by \cite{hansen2021opinion}, in the limit this diffusion process achieves $\ueF x_u = \veF x_v$ by adjusting the restriction maps to the signal. However, this might be still limiting for modelling complex interactions, as one node's private opinions cannot change at all. 

\subsection{Joint Opinion-Expression Diffusion}

To overcome the unrealistic rigidity of opinions of the previous model, the work of \cite{hansen2021opinion} proposes a novel \textit{joint opinion-expression diffusion} setting that puts together both the \textit{Learning to Lie} and the Regular Sheaf Diffusion processes.
From a theoretical perspective, this results in the following system of non-linear ODEs: 
\begin{equation} 
\label{eq:JointDiff}
\begin{dcases}
    \frac{d}{dt}\F^*(t) = -\beta \Delta_{X(t)}\F^*(t) \\
    \frac{d}{dt}X(t) = -\alpha \Delta_{\F(t)}X(t)
\end{dcases}
\end{equation}
where $\alpha, \beta \in \R$ are coefficients that control the diffusion strength of each ODE, prioritising the diffusion either over the restriction maps or over the features.

The following lemma can help us understand the inductive bias brought by Equation \eqref{eq:JointDiff}:
\begin{lemma}
    \cite{hansen2021opinion} Let's consider $\Psi(X(t),\F(t)) = X(t)^T{\delta (t)}^T\delta (t) X(t)$ where $X(t), \F(t)$ is a solution of \eqref{eq:JointDiff}. Then $\Psi(X(t),\F(t)) \geq 0$ and $\frac{d\Psi(X(t),\F(t))}{dt} \leq 0$ with zero attained in both if and only if $\ueF x_u = \veF x_v$ for all $u,v \unlhd e$.
\end{lemma}
This means that we may see this process as gradient descent on the sheaf Dirichlet energy \cite{bodnar2022neural}, which means we are evolving towards a sheaf equilibrium leveraging both the node's features and the restriction maps. Using LaSalle's invariance principle \cite{hansen2021opinion} we can prove that we actually converge to an equilibrium point with $\ueF x_u = \veF x_v$ for all $u,v \unlhd e$.

Even though we have ensured convergence to some point $(x_\infty,\delta_\infty)$, it is still necessary to check when $x_\infty \neq 0$. This is due to the fact that, if the trajectories of (\ref{eq:JointDiff}) all converged to zero, this system would also oversmooth the signal $X(t)$. To do this, we resort to the following result:
\begin{theorem}
\cite{hansen2021opinion}
    If the initial conditions of the system (\ref{eq:JointDiff}) are $\delta_0, x_0$ and one of the diagonal blocks of $\alpha \delta_0^T \delta_0 -  \beta x_0 x_0^T$ fails to be semidefinite, the trajectory of (\ref{eq:JointDiff}) converges to a point $(x_\infty, \delta_\infty)$ with $x_\infty \neq 0$.
\end{theorem}
This shows a clear advantage concerning regular sheaf diffusion, as we can easily check if the system will converge to a non-zero solution. In addition to this, we also have the following result:
\begin{corollary}
\cite{hansen2021opinion}
    If $(x_0,\delta_0)$ are some initial conditions, then there exists a $k$ such that the trajectory of (\ref{eq:JointDiff}) with initial conditions $(kx_0,\delta_0)$ converges to a point $(x_\infty,\delta_\infty)$ with $x_\infty \neq 0$.
\end{corollary}
Therefore, we can always guarantee the existence of non-zero solutions by just tuning one parameter. Note, however, that this is not the case for regular sheaf diffusion, where orthogonal restriction maps are necessary to get a similar guarantee. Another advantage is that the diffusion leverages both restriction maps and the node's features to reach a point of equilibrium. This is particularly useful in cases of heterophily, as this process may shut off communication among nodes with different opinions without the need of a learnable function.

\begin{figure}
    \centering
    \includegraphics[scale=0.25]{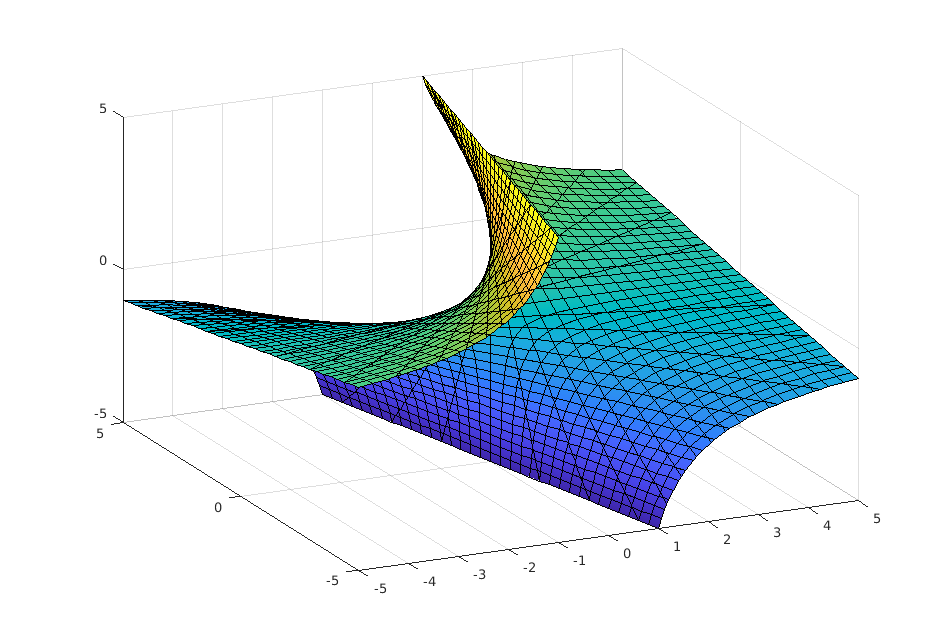}
    \caption{A visualization of the solutions of $zx = ty$, with $t=1$ because the ones with $t=0$ are trivially $t=0$ and $y=0$. This is, in fact, an affine projection of a projective variety. This corresponds with the set of equilibrium points of Eqs. (\ref{eq:JointDiff}) in the case of 1-dimensional stalks and a graph with only 2 adjacent nodes, so $z,t$ would be the restriction maps while $x,y$ would be the node's features. Sheaf diffusion only evolves $x,y$, so it may miss close points of equilibrium in the $z$ axis.}
    \label{fig:EquilibriSheaf}
\end{figure}

\framedtext{
    \textit{Example.} Figure \ref{fig:EquilibriSheaf} let us visualize the usefulness of diffusing both restriction maps and node's features by depicting the points of equilibrium of Eqs.~\ref{eq:JointDiff} for a particular case example. On the one hand, the trajectory used in the Joint Diffusion has the freedom to move in any direction to reach a point of equilibrium. On the other hand, the trajectory used in classical Sheaf Diffusion (Equation~\ref{eq:SheafDiffusion}) can only move in a horizontal plane, which means we may miss close points of equilibrium in the vertical axis.
}

\subsection{Joint Diffusion Sheaf Neural Networks}

With the previous dynamical system \eqref{eq:JointDiff} in mind, we propose a new SNN model as the natural conclusion of this section:

\begin{definition}
With the setting described in the Definition \ref{def:SNN} of SNNs, we define the $t$ layer of a Joint diffusion Sheaf Neural Network (JdSNN) through the following equations:
\begin{equation}
\label{eq:JdSNN}
\begin{dcases}
X(t+1) =  X(t) - \\ 
\quad \sigma((\textbf{I}_{nd} - \alpha\Delta_\F(t))(I_n \otimes W_1(t)) X W_2(t)) \\
\F^*(t+1) = \F^*(t) - \\
\quad \sigma((\textbf{I}_{2md} - \beta\Delta_{X(t)})(I_n \otimes W^*_1(t)) \F^* W^*_2(t))  
\end{dcases}
\end{equation}

where $\alpha,\beta$ are weights to give more importance to one of the diffusions depending on the problem.
\footnote{For example, if we know the dataset is heterophillic we may want to set $\beta > \alpha$ to prioritize the diffusion of the restriction maps, potentially preventing oversmoothing.}
\end{definition}

This new variant has an extra inductive bias towards learning on heterophilic data, as it may prevent oversmoothing of the signal $x_u$ by "oversmoothing" the restriction maps $\ueF$ instead. For instance, let us consider the case of two adjacent nodes $u,v$ of different classes and significantly different features; whereas the second equation of \eqref{eq:JdSNN} would make the restriction maps evolve to avoid mixing the features of $u$ and $v$ too much, it is unclear if --and how-- the features from $u,v$ would mix when using $\ueF = MLP(x_u||x_v)$. Moreover, all of this is achieved with $2d^2$ parameters per diffusion layer coming from $W_1^*,W_2^*$, instead of the at least $2d^2c$ ones of usual SNNs (with $d$ being the stalk dimension and $c$ the number of original features and $2d^2c$ is the size of the weights matrix used to predict the restriction maps). Here it is important to note that the number of parameters in JdSNNs no longer scale with $c$, making this model suitable for contexts where extensive parameterizations may not be possible. Examples of this are small datasets or some scenarios of federated learning where the size of the features makes linear layers impossible to apply \cite{gabrielli2023survey}.

However, this new model also presents some setbacks, as 
it is not possible to impose some constraints such as orthogonal or diagonal restriction maps, which have proved to be useful when training SNNs \cite{bodnar2022neural,duta2024sheaf}. We also lose the universal sheaf approximation result presented by \cite{bodnar2022neural}, but this actually represents a trade-off between adding an inductive bias for heterophilic data instead of using the most general model.

\subsection{Rotation Invariant Sheaf Neural Networks}

As we mentioned in the previous section, JdSNNs gain an inductive bias at the cost of using more complex dynamics. In hopes of simplifying the dynamics while keeping the inductive bias, we introduce another alternative that explores a new way of learning the restriction maps within the usual SNN formulation of Definition~\ref{def:SNN}. 

In our proposal, instead of using $\ueF(t+1) = MLP(x_u(t) || x_v(t))$, we consider the vector $x_e(t) = \ueF(t) x_u(t) - \veF(t) x_v(t)$ --that represents the conversation in the edge space--, and set $\ueF(t+1) = MLP(x_e(t)x_u(t)^T)$. With this approach, the number of parameters we use is $d^4$, which is more than JdSNNs but still does not scale with the total number of features $c$. In this model, instead of learning the restriction maps from all the features, we are assuming that they only depend on the relationship between the node's private opinion $x_u$ and the public conversation $x_e$:
$$x_e x_u^T = \begin{pmatrix}
<x_{e1},{x_u}_1> & ... & <x_{e1},{x_u}_d> \\
... & ... & ... \\
<x_{ed},{x_u}_1> & ... & <x_{ed},{x_u}_d>
\end{pmatrix},$$

where $x_{ei}, x_{uj}$ are the i-th and j-th row vectors of $x_e,x_u$. With this in mind, the following proposition can be easily proven:

\begin{proposition} \label{prop:rotation_invariant}
The restriction maps of a SNN learnt using $\ueF(t+1) = MLP((\ueF(t) x_u - \veF(t) x_v)x_u^T)$ are feature-wise rotation invariant w.r.t $X$. That is, given $Q$ an orthogonal matrix, then the restriction maps computed for $X$ and $QX$ will be equal.
    
\textit{Proof in Appendix \ref{appendix:proof_rotation_invariant}.}
\end{proposition}


Due to this result, we call this SNN variant \textbf{Rotation invariant Sheaf Neural Networks} (RiSNN). Please note that only the restriction maps are rotation invariant. However, it is possible to obtain a GNN with rotation equivariance by setting $\sigma = id$ and $W_2=Id$. A more intuitive way of thinking about this model would be performing an attention-like mechanism on multiple cosine-similarities.

\section{Benchmark Evaluation} \label{section:benchmark}

In this section we test the introduced opinion-inspired SNN models under the benchmark defined in \cite{bodnar2022neural}.
\footnote{Except the film dataset, as equivalent results to \cite{bodnar2022neural} can be obtained ignoring the graph structure.}

\begin{table*}[]
\centering
\caption{Accuracy obtained by the considered models on the node classification benchmark originally used by \cite{bodnar2022neural}, sorted by homophily degree. Top three models are coloured by {\color[HTML]{FF0000} First}, {\color[HTML]{0000FF} Second}, {\color[HTML]{9900FF} Third}. Our models are: RiSNN-NoT, RiSNN, JdSNN, and JdSNN-NoW.}
\begin{adjustbox}{width=1\textwidth}
\begin{tabular}{lcccccccc}
\hline
\multicolumn{1}{c}{
\textbf{Dataset}}     & \textbf{Texas}                    & \textbf{Wisconsin}                & \textbf{Squirrel}                 & \textbf{Chameleon}                & \textbf{Cornell}                  & \textbf{Citeseer}                 & \textbf{Pubmed}                   & \textbf{Cora}                     \\
\hline
\multicolumn{1}{c}{\textbf{\#Nodes}}     & \textbf{183}                      & \textbf{251}                      & \textbf{5201}                     & \textbf{2277}                     & \textbf{183}                      & \textbf{3327}                     & \textbf{18717}                    & \textbf{2708}                     \\
\multicolumn{1}{c}{\textbf{\#Edges}}     & \textbf{295}                      & \textbf{466}                      & \textbf{198493}                   & \textbf{31421}                    & \textbf{280}                      & \textbf{4676}                     & \textbf{44327}                    & \textbf{5278}                     \\
\multicolumn{1}{c}{\textbf{\#Homophily}} & \textbf{0.11}                     & \textbf{0.21}                     & \textbf{0.22}                     & \textbf{0.23}                     & \textbf{0.3}                      & \textbf{0.74}                     & \textbf{0.80}                     & \textbf{0.81}                     \\
\hline
\textbf{RiSNN-NoT}                       & {\color[HTML]{FF0000} 87.89±4.28} & 88.04±2.39                        & 51.24±1.71                        & {\color[HTML]{0000FF} 66.58±1.81} & 82.97±6.17                        & 75.07±1.56                        & 87.91±0.55                        & 85.86±1.31                        \\
\textbf{RiSNN}                         & 86.84±3.72                        & 87.84±2.60                      & {\color[HTML]{FF00FF} 53.30±3.30} & 65.15±2.40                        & 85.95±6.14                        & {\color[HTML]{FF00FF} 76.23±1.81} & 88.00±0.42                        & 85.27±1.11                        \\
\hline
\textbf{JdSNN-NoW}                       & 87.30±4.53                        & 88.43±2.83                        & 51.28±1.80                        & 66.45±3.46                        & 84.59±6.95                        & 75.93±1.41                        & 88.09±0.49                        & 84.39±1.47                        \\
\textbf{JdSNN}                         & {\color[HTML]{FF00FF} 87.37±5.10} & {\color[HTML]{0000FF} 89.22±3.42} & 49.89±1.71                        & 66.40±2.33                        & 85.41±4.55                        & 73.27±1.86                        & 88.19±0.55                        & 85.43±1.73                        \\
\hline
\textbf{Conn-NSD}                        & 86.16±2.24                        & 88.73±4.47                        & 45.19±1.57                        & 65.21±2.04                        & 85.95±7.72                        & 75.61±1.93                        & 89.28±0.38                        & 83.74±2.19                        \\
\textbf{Best-NSD}                        & 85.95±5.51                        & {\color[HTML]{FF0000} 89.41±4.74} & {\color[HTML]{FF0000} 56.34±1.32} & {\color[HTML]{FF0000} 68.68±1.73} & {\color[HTML]{FF00FF} 86.49±7.35} & {\color[HTML]{FF0000} 77.14±1.85} & {\color[HTML]{0000FF} 89.49±0.40} & {\color[HTML]{0000FF} 87.30±1.15} \\
\textbf{Best-NSP}                        & 87.03±5.51                        & {\color[HTML]{FF00FF} 89.02±3.84} & 50.11±2.03                        & 62.85±1.98                        & 76.49±5.28                        & {\color[HTML]{0000FF} 76.85±1.48} & {\color[HTML]{FF00FF} 89.42±0.33} & {\color[HTML]{FF0000} 87.38±1.14} \\
\textbf{Best-BC-NLSD}                    & {\color[HTML]{0000FF} 87.57±5.43} & {\color[HTML]{FF0000} 89.41±2.66} & {\color[HTML]{0000FF} 54.62±2.82} & {\color[HTML]{FF00FF} 66.54±1.05} & {\color[HTML]{FF0000} 87.30±6.74} & 76.03±1.56                        & 89.39±0.43                        & {\color[HTML]{FF00FF} 87.00±1.23} \\
\textbf{Best-MLP-NLSD}                   & 86.22±3.91                        & {\color[HTML]{FF00FF} 89.02±3.19} & 51.96±2.65                        & 65.37±2.73                        & {\color[HTML]{0000FF} 87.03±4.49} & 76.11±1.81                        & {\color[HTML]{FF0000} 89.60±0.29} & 86.38±1.20
\\
\hline
\end{tabular}
\end{adjustbox}
\label{table:benchmarksres}
\end{table*}
\paragraph{Ablation Study}
For this evaluation, we consider a total of 4 different versions of our models: RiSNN and JdSNN are the ones we have already discussed. RiSNN with no time dependency (RiSNN-NoT) is a reinterpretation of the regular RiSNN by setting $x_e(t) = x_u(t)-x_v(t)$; the goal of this modification is to stabilize the gradient. JdSNN with no learnable weights (JdSNN-NoW) is a simplification of our full version where the restriction maps are computed without any parameters --i.e. obtained by removing the non-linearity, considering $W_1^*, W_2^* = Id$, and setting $\ueF(0) = Id$ for all $u \in V, e \in E$. These 2 simpler variants are shown to achieve equivalent results in some cases.

Table \ref{table:benchmarksres} shows the results of our models against the best version of the architectures proposed in \cite{bodnar2022neural,suk2022surf,barbero2022sheaf,zaghen2024nonlinear}. As we can see in Table \ref{table:benchmarksres}, we mostly get statistically equivalent results to those of other methods while inferring the sheaf structure with significantly less parameters. Consequently, this shows that the inductive bias we have traded for learnable parameters is useful and does not negatively affect performance when applied to heterophilic datasets. 

We note, though, that our models --especially JdSNN-- do not perform greatly in two heterophilic datasets (Squirrel and Chameleon). However, they also happen to be the biggest ones, so overall this still aligns with what we have discussed so far: on the one hand, high amounts of data allow for a complex $MLP(x_u||x_v)$ that approximates $\ueF$ accurately. On the other hand, when there is not enough data for a good approximation to be learned, the inductive bias we provide is capable of replacing these extensive parametrizations.

However, previous work \cite{platonov2024characterizing} highlights the limitations of these benchmarks for properly evaluating the model's abilities to capture heterophilic interactions. In fact, there is a pressing need in the graph community to find and/or design new datasets that can help to properly assess current model architectures, providing with meaningful insights about their actual capabilities in different settings.  

\section{Synthetic Evaluation} \label{sect:SynthExp}

Arguably, the need of meaningful datasets to test SNNs' capabilities is even more crucial than in the broader graph domain given its recent formalization and its particular geometric intricacies. However, apart from evaluating them on graph benchmarks, previous sheaf-based works only assess their proposed models considering either denoising tasks~\cite{zaghen2024nonlinear,bodnar2022neural}, or heterophilic scenarios with unrealistic single-feature $d=1$ setups~\cite{hansen2020sheaf,bodnar2022neural}.\footnote{By construction our model relies on $d>1$, the most usual case in real-world scenarios.} 

Given this context, this section presents a novel method to produce synthetic datasets, which we then leverage to evaluate different aspects and capabilities of SNNs --and those of our introduced variants. More specifically, the proposed pipeline is divided into two main processes --feature and edge generation--, and is overall designed to address the following research questions:
\begin{enumerate}
    \item How does node features' noise affect performance?
    \item What is the impact of the ratio of intra-class edges vs inter-class edges?
    \item How does the amount of data available correlate with performance for different SNN variants?
\end{enumerate}



\subsection{Synthetic Features Generation}
As we previously mentioned, we aim to measure the impact of noise in our models as well as the impact of heterophily and the amount of data available. This means that generating classes by adding noise to different expected values is not a viable option, as we would only be able to answer the first question. Consequently, we turn to manifold sampling to generate complex patterns for the different classes.

\begin{figure}
\centering
\includegraphics[scale=0.4]{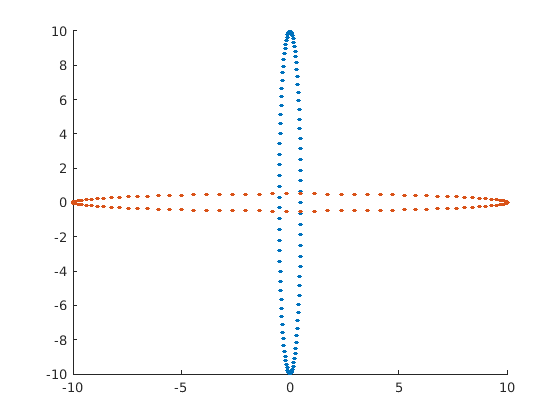}
   \caption{Graphic plot of two classes, red and blue, generated by sampling a 2-dimensional ellipsoid's surface. In this representation, it's easy to see how they're not linearly separable but distinct, and that the expected value for both of them is zero.}
\label{fig:Ellipsoid}
\end{figure}

In particular, we choose the surface of different $n$-dimensional ellipsoids for each class, all sharing one central point as can be seen in Fig.~\ref{fig:Ellipsoid}. The reason behind this choice relies on their symmetrical properties, producing classes that are not linearly separable while at the same time, they all share the same expected value. This second condition is key as it ensures that non-trivial aggregation will be necessary.

\subsection{Synthetic Edge Generation}
Regarding the edge generation, we use a variation of the Watts-Strogatz model \cite{watts1998collective} that also introduces inter-class correlations. 
But before describing the pipeline, let us first introduce the involved notation: $N$ denotes the number of nodes, $K$ the desired degree of each node (should be an even number), $p \in [0,1]$ a probability-based parameter, $n_c$ the number of node classes, and $R^c \in \mathcal{M}_{n_c \times n_c}$ the matrix such that $R^c_{ij} = Pr(\{node_1,node_2\} \in E | node_1 \in class_i, node_2 \in class_j)$. We also consider a $het$ coefficient representing the level of heterophily in the graph, which is defined as the probability that a node is connected to another class. This in turn allows us to write the matrix of inter-class correlations as:
\begin{equation*}
    R^c = \left(1-\frac{n_c+1}{n_c-1}het\right)Id + \frac{het}{n_c-1}*\boldsymbol{1_{n_c}}\boldsymbol{1_{n_c}}^T,
\end{equation*}
with $1-het$ in the diagonal entries, and $\frac{het}{n_c-1}$ in the rest.

Bearing this notation in mind, the steps to generate the final graph connectivity of the data are as follows:
\begin{enumerate}
    \item Label the nodes from $0$ to $N-1$ and assign each node a class uniformly at random.
    \item Construct a regular ring lattice. That is $(i,j) \in E \Leftrightarrow 0 < |i-j| mod (N-1-K/2) \leq K/2$.
    \item For each node $i=0,...,N-1$ we consider its  $K/2$ rightmost edges and with probability $p$ rewire it to another node. To choose the node we first sample a class from the distribution of the $i$-th row of $R^c$, and then we choose uniformly at random a node of said class that is not connected to node $i$.
\end{enumerate}

\subsection{Experimental setup}

With the focus on assessing how the restriction maps are learned, we set $W_1 = Id, W_2 = Id$, and $\sigma=Id$ in the setting introduced in Definition \ref{def:SNN}, as done in the work \cite{bodnar2022neural}.


Analogously to our previous Benchmark Evaluation, in these experiments we also test the full JdSNN and RiSNN models, as well as their respective simpler variants JdSNNnoW and RiSNN-NoT (please see Section \ref{section:benchmark}). Additionally, here we also consider another joint diffusion variation, JdSNN-W0, which initializes the restriction maps with a learnable function --$\F^*(0) = MLP(x_u||x_v)$-- but does the Dual Diffusion without any parameters --$\F^*(t+1) = \F^*(t) - \Delta_{X(t)}\F^*(t)$.






We compare these models against the following baselines: VanillaSheaf, that is $\ueF = Id$ for all $u \in V, e \in E$; a regular MLP with one layer and $id$ as activation function; and a SNN with diagonal restriction mappings.

\begin{remark}
    For the Joint Diffusion based models, we wish to test the separation power of each of the diffusion processes. To this end, in the experiments involving JdSNNs, only 1 feature channel and 3-dimensional stalks are considered --i.e. the hidden features have shape $3 \times 1$. Conversely, for RiSNNs evaluations we use 3-dimensional stalks and 5 feature channels to fully utilize the scalar products in $x_ex_u^T$.
\end{remark}

\subsection{Results}
This section describes the main takeaways of our extensive evaluation of the generated synthetic datasets, with the focus on answering the fundamental research questions raised at the beginning of this section. 
Apart from Figure~\ref{fig:LineNoise} above, we also include in Appendix \ref{appendix:synthexpRes} the plots associated with the rest of performed experiments (Figures \ref{fig:JdSNNFeatNoise},  \ref{fig:JdSNNHet} and \ref{fig:JdSNNData}).

\begin{figure}[]
\includegraphics[scale = 0.35]{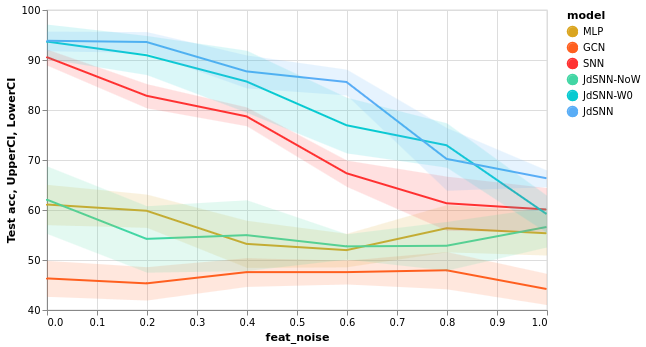}
   \caption{JdSNNs variants' accuracy results with a 95\% confidence interval when increasing the percentage of feature Gaussian noise in the data. We may observe how our methods are more robust to noise than regular SNNs. We can also observe how GCNs and MLPs underfit the data.}
\label{fig:LineNoise}
\end{figure}

\paragraph{Joint diffusion without any parameters} Overall, the JdSNN version without parameters, JdSNN-noW, has an advantage over GCNs (Figures \ref{fig:JdSNNFeatNoise},  \ref{fig:JdSNNHet} and \ref{fig:JdSNNData}). However, in our task one layer oversmooths the representation, so setting $\F(0) = Id$ is a bottleneck to the model's performance.

\paragraph{The inductive bias towards heterophily} Our proposed models, and specially the Joint Diffusion variants, show a clear bias towards heterophilic data as seen in Figure \ref{fig:JdSNNHet}. When increasing the number of inter-class edges the difficulty of the problem also increases, as more different types of interactions must be modelled. In this aspect, the performance of standard SNNs decreases significantly more than JdSNNs.

\paragraph{Joint Diffusion in cases of noisy data} JdSNNs are more robust to noise in the cases of small dimensionality of the data. As one may observe in Fig.~\ref{fig:LineNoise}, the version using only a learnable initialization and the full model are significantly better than a regular SNN when adding noise, up to the point where the features are 70\% noise.

\paragraph{Rotation invariant SNNs in cases of noisy data} RiSNNs in general are more sensitive to noisy features than SNNs as seen in Figure \ref{fig:JdSNNFeatNoise}. Nonetheless, the RiSNN-NoTime version is slightly better than the original one. In cases of non-noisy data, the original formulation of RiSNNs tends to have a slight advantage over the version that doesn't use the restriction maps of the previous layer.

\paragraph{Amount of available data and the performance of SNNs} SNNs' performance is greatly increased in problems with a large number of edges as seen in Figure \ref{fig:JdSNNData}. This allows the MLP which computes the restriction maps to get a better approximation. The inductive bias introduce by our variants, JdSNN and RiSNN, restricts the possible relationships between nodes, so they have an advantage when the sheaf structure cannot be accurately computed \ref{fig:JdSNNData}.

\section{Conclusions}
This work proposes and evaluates new ways of inferring the sheaf structure of a graph taking inspiration from complex opinion dynamics processes. Through these, we show how using opinion dynamics interpretations of sheaves may lead to a more intuitive understanding of SNNs and how they operate. When benchmarked, our new proposed models present a clear inductive bias towards heterophily, and they are able to achieve equivalent performance to regular SNNs while using fewer parameters to infer the sheaf structure. In particular, this reduction of the number of parameters opens the door to applying sheaf-based methods in domains where extensive parameterizations cannot be used. Furthermore, the geometrical properties of the introduced RiSNN model may enable the application of sheaves to geometrical deep learning problems.

Last but not least, this paper also proposes a novel way of evaluating SNNs \textit{(i)} using the surface of $n$-dimensional ellipsoids to generate features, and \textit{(ii)} generating graph connectivities with different degrees of heterophily. 
From extensive experimentation on generated datasets, we observe that SNNs benefit from having graphs with lots of edges, while our introduced variants take the lead in graphs with low connectivity. 

Overall, we reckon that our introduced models and synthetic data pipelines can pave the way for interesting future research in this exciting area, fostering a deeper understanding of the inner workings and capabilities of SNNs.\footnote{A further discussion of limitations and future work can be found in Appendices \ref{appendix:limitations} and \ref{appendix:future_work}, respectively.}

\section*{Acknowledgments}
The authors would like to thank the Private Foundation Pere Mir-Puig and CFIS for the funding provided to Ferran Hernandez to facilitate his stay at the University of Cambridge.

\bibliography{example_paper}
\bibliographystyle{icml2024}

\newpage
\appendix
\onecolumn

\section{Proof of Proposition \ref{prop:GrafSmooth}} \label{appendix:proof_oversmoothing}
For this proof, we use that $lim_{t \rightarrow \infty} X(t)$ is the orthogonal projection of $X(t)$ to $ker(L)$. So, we will prove that $ker(L)$ only contains vectors constant in all connected components. Given $v \in ker(L)$, we compute $Lv$:
\begin{align}
\label{deltaV}
    L v &= 
    \begin{bmatrix}
           \sum\limits_{j | (1,j) \in E} (v_1 - v_j) \\
           \sum\limits_{j | (2,j) \in E} (v_2 - v_j) \\
           \vdots \\
           \sum\limits_{j | (n,j) \in E} (v_n - v_j)
         \end{bmatrix}
  \end{align}
As $v \in ker(L)$, $Lv = 0$, so $v^tLv = 0$. Expanding $v^tLv$ yields
$$v^tL v = \sum\limits_{i = 1}^n \sum\limits_{j|(i,j)\in E} v_i(v_i-v_j) $$
Now, if we select $i < j, (i,j) \in E$, the sum contains the terms $v_i(v_i-v_j)$ and $v_j(v_j-v_i)$ exactly once. So, by grouping terms we get $(v_i-v_j)^2$. This directly implies: $v^tL v = \sum\limits_{i < j, (i,j) \in E} (v_i - v_j)^2$. This and this quantity is zero if and only if $v_i = v_j \forall \{i,j\} \in E$, which concludes our proof. \hfill $\blacksquare$

\section{Proof of Proposition \ref{prop:rotation_invariant}} \label{appendix:proof_rotation_invariant} 

Let's suppose $X' = XQ$ with Q an orthogonal matrix, we proceed by induction on the number of layers. Since $t=0$ is not defined, we set it to a constant ($Id$) the rotation trivially does not affect this. Then if $\ueF'(t) = \ueF(t)$
\begin{equation*}
    \begin{split}
    \ueF'(t+1) &= MLP((\ueF'(t) x_u' - \veF'(t) x_v')x_u'^T)   \\ &=MLP((\ueF'(t) x_u'Q - \veF'(t) x_v'Q)Q^Tx_u'^T)  \\ 
    &=MLP((\ueF(t) x_uQ - \veF(t) x_v)x_u^T) \\
    &= \ueF(t+1) 
\end{split}
\end{equation*}

Where we have used both our induction hypothesis and $QQ^T = Id$. \hfill $\blacksquare$

\section{Synthetic Experiments Results}
\label{appendix:synthexpRes}
\begin{figure*}[ht]
\begin{center}
\includegraphics[scale = 0.37]{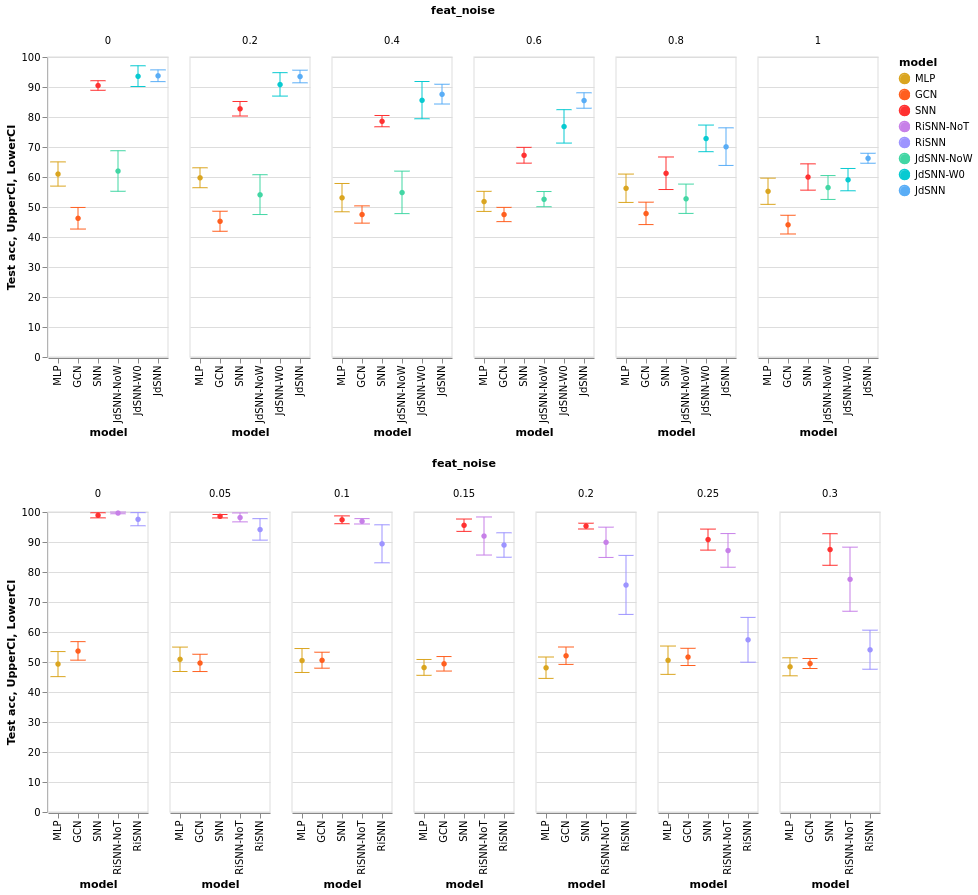}
\end{center}
   \caption{JdSNNs variants' results at the top and RiSNNs variants' results at the bottom when increasing feature Gaussian noise in the data}
\label{fig:JdSNNFeatNoise}
\end{figure*}

\begin{figure*}[ht]
\begin{center}
\includegraphics[scale = 0.27]{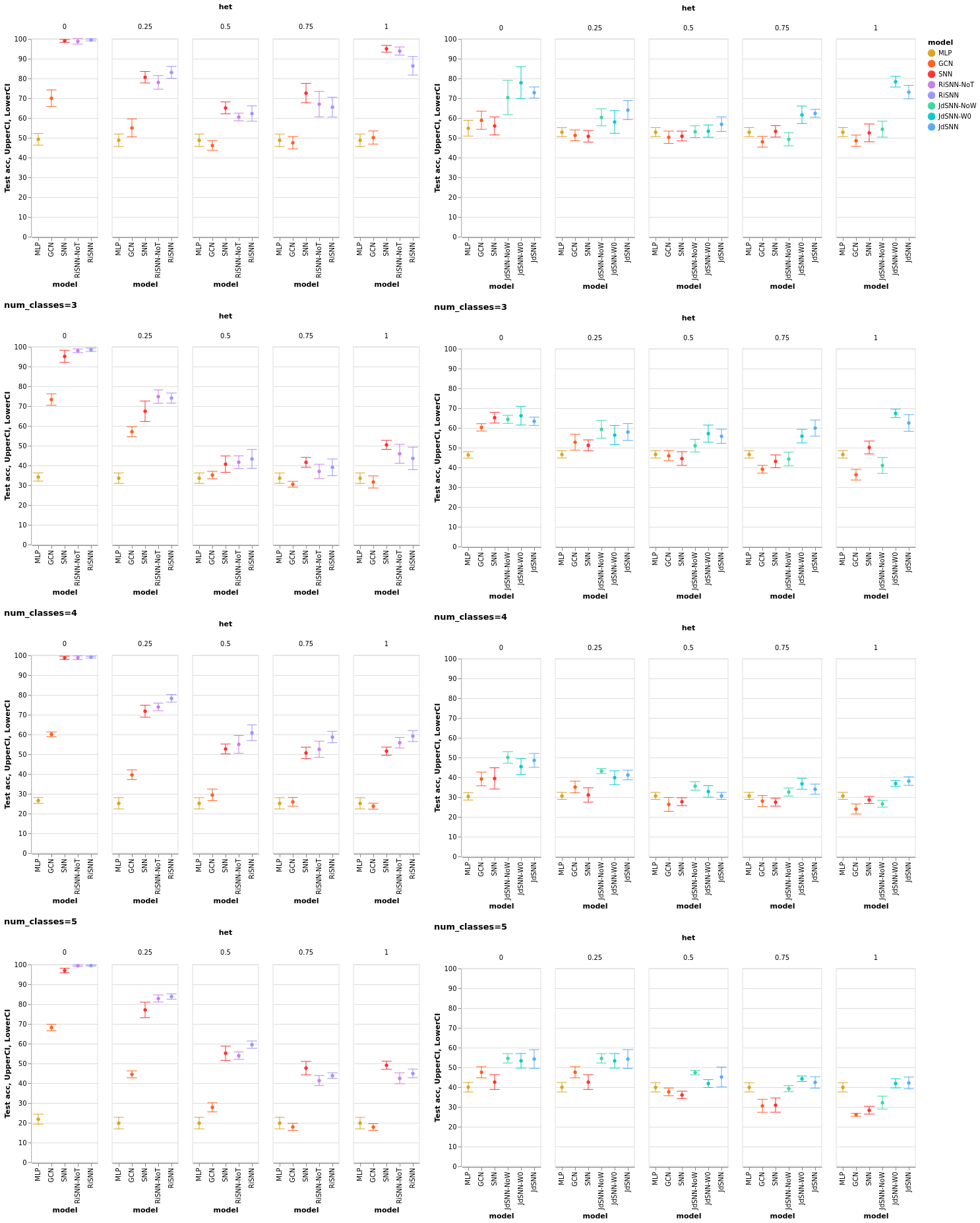}
\end{center}
   \caption{RiSNNs variants' results on the right and JdSNNs variants' results on the right when increasing the number of classes and the heterophily coefficient of the data}
\label{fig:JdSNNHet}
\end{figure*}

\begin{figure*}[ht]
\begin{center}
\includegraphics[scale = 0.36]{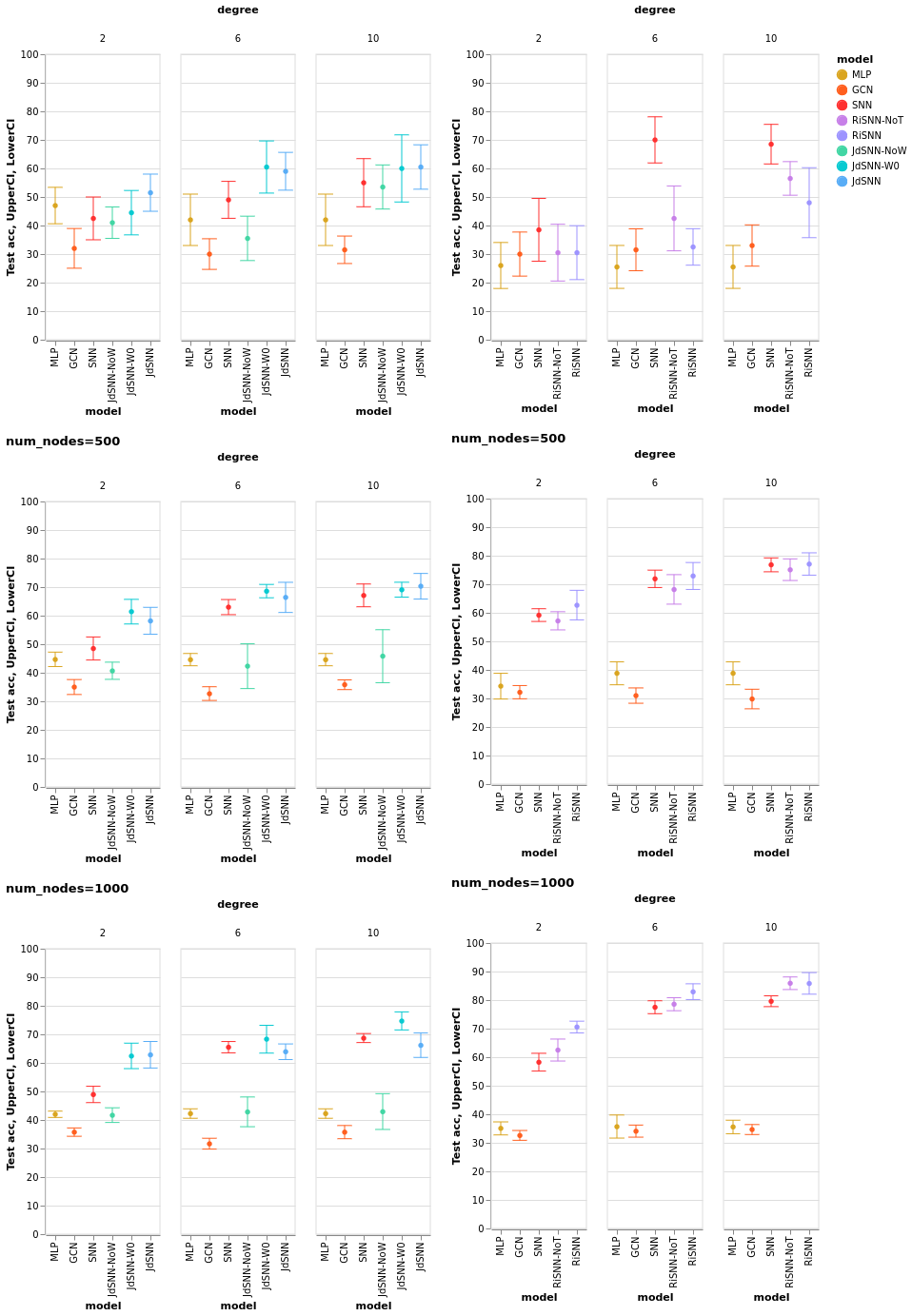}
\end{center}
   \caption{JdSNNs variants' results on the left and RiSNNs variants' results on the right when increasing the amount of nodes and edges of the data.}
\label{fig:JdSNNData}
\end{figure*}

In this section, we can observe the results which answer the three questions presented in Section \ref{sect:SynthExp}:
\begin{itemize}
    \item Figure \ref{fig:JdSNNFeatNoise}: How does noise impact the performance of SNNs and their variants?
    \item Figure \ref{fig:JdSNNHet}: How does the ratio of intra-class edges vs inter-class edges impact the performance of SNNs and their variants? 
    \item Figure \ref{fig:JdSNNData}: How does the amount of data available impact the performance of SNNs and their variants?
\end{itemize}

\clearpage

\clearpage

\section{Limitations} \label{appendix:limitations}
The main limitation of the models proposed is that they are more complex to backpropagate; for example, using the restriction maps to define the normalization matrix $D$ as in \cite{bodnar2022neural} can lead to exploding or undefined gradients. For this reason, it is necessary to detach $D$ from the backpropagation algorithm.

Another limitation is that they are usually slower than SNNs. As was shown in \cite{bodnar2022neural}, the time complexity of an SNN is $\mathcal{O}(nc^2+mc)$ if the restriction maps are diagonal and $\mathcal{O}(n(c^2+d^3)+m(c+d^3))$ if they're general, where $n$ is the number of nodes, $m$ the number of edges, $c$ the total number of features, and $d$ the stalk dimension. When it comes to our variants, JdSNNs and RiSNNs both have a time complexity of $\mathcal{O}(n(c^2+d^3)+md(c+d^3))$. This implies that using these models with large $d$ or $m$ will lead to noticeably slower execution times. Nonetheless, we believe there is room for improvement in this aspect.

\section{Future Work} \label{appendix:future_work}
SNNss are a relatively new GNN \cite{hansen2020sheaf,bodnar2022neural}, so there are many interesting research directions. In this section, we highlight a couple of them that we consider particularly interesting. The first one is related to federated learning while the latter with geometric deep learning. In federated learning, there have been approaches which are graph-based \cite{gabrielli2023survey}. In this context, there are cases with millions of features, consequently, using regular MLPs on those features is not viable as a basic linear layer would have too many parameters. This implies that regular SNNs may not be used, but the new variants that have been proposed, RiSNN and JdSNN, have a number of parameters which only scales with the stalk dimension $d$ which means sheaves may be used directly in federated learning. In some geometric problems on graphs, some extra properties regarding rotation are necessary \cite{duval2023hitchhiker}. While this might prevent from implementing regular SNNs, RiSNNs might be more easily adaptable to be used in said problems.

\end{document}